# Building Footprint Generation Using Improved Generative Adversarial Networks

Yilei Shi, *Member, IEEE*, Qingyu Li, Xiao Xiang Zhu, *Senior Member, IEEE*

*Abstract*—Building footprint information is an essential ingredient for 3-D reconstruction of urban models. The automatic generation of building footprints from satellite images presents a considerable challenge due to the complexity of building shapes. In this work, we have proposed improved generative adversarial networks (GANs) for the automatic generation of building footprints from satellite images. We used a conditional GAN with a cost function derived from the Wasserstein distance and added a gradient penalty term. The achieved results indicated that the proposed method can significantly improve the quality of building footprint generation compared to conditional generative adversarial networks, the U-Net, and other networks. In addition, our method nearly removes all hyperparameters tuning.

## I. INTRODUCTION

**B**UILDING footprint generation is of great importance to urban planning and monitoring, land use analysis, and disaster management. High-resolution satellite imagery, which can provide more abundant detailed ground information, has become a major data source for building footprint generation. Due to the variety and complexity of buildings, building footprint requires significant time and high costs to generate manually (see Fig. 1). As a result, the automatic generation of a building footprint not only minimizes the human role in producing large-scale maps but also greatly reduces time and costs.

Previous studies focusing on building footprint generation can be categorized into four aspects: (a) edge-based, (b) region-based, (c) index-based, and (d) classification-based methods. In edge-based methods, regular shape and line segments of buildings are used as the most distinguishable

This work is supported by the European Research Council (ERC) under the European Union's Horizon 2020 research and innovation programme (grant agreement no. ERC-2016-StG-714087, acronym: So2Sat, www.so2sat.eu), the Helmholtz Association under the framework of the Young Investigators Group "SiPEO" (VH-NG-1018, www.sipeo.bgu.tum.de), Munich Aerospace e.V. Fakultät für Luft- und Raumfahrt, and the Bavaria California Technology Center (Project: Large-Scale Problems in Earth Observation). The authors thank Planet provide the datasets. (*Corresponding Author: Xiao Xiang Zhu*)

Y. Shi is with the Chair of Remote Sensing Technology (LMF), Technische Universität München (TUM), 80333 Munich, Germany (e-mail: yilei.shi@tum.de)

Q. Li is with Signal Processing in Earth Observation (SIPEO), Technische Universität München (TUM), 80333 Munich, Germany (e-mail: qingyu.li@tum.de)

X.X. Zhu with the Remote Sensing Technology Institute (IMF), German Aerospace Center (DLR) and Signal Processing in Earth Observation (SIPEO), Technische Universität München (TUM), 80333 Munich, Germany (e-mail: xiaoxiang.zhu@dlr.de)

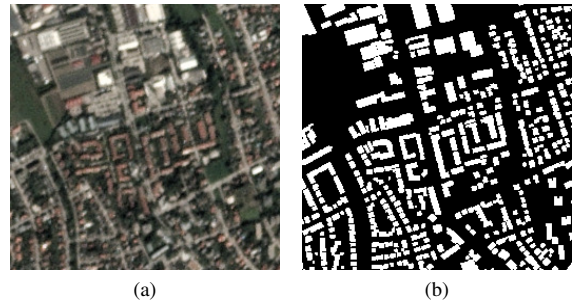

Fig. 1. (a) Optical satellite imagery of PlanetScope; (b) Building footprint from OpenStreetMap

features for recognition [1]. Region-based methods identify building regions through image segmentation [2]. For index-based methods, a number of building feature indices are used to describe the characteristics of buildings, which indicate the possible presence of buildings [3]. Classification-based methods, which combine spectral information with spatial features, are among the most widely used approaches, since they can provide more stable and generalized results than the other three methods.

Traditional classification-based methods consist of two steps: feature extraction and classification. Among them, the support vector machine (SVM) and random forest (RF) are two popular classification approaches in the remote sensing (RS) domain. However, an SVM will consume too many resources when used for big data applications and large area classification problems, and multiple features should be engineered to feed the RF classifier for efficient use. Recent advances in traditional classification methods, e.g. [4] and [5], show promising results.

Over the past few years, the most popular and efficient classification approach has been deep learning (DL) [6], which has the computational capability for big data. DL methods combine feature extraction and classification and are based on the use of multiple processing layers to learn good feature representation automatically from the input data. Therefore, DL usually possesses better generalization capability, compared to other classification-based methods. In terms of particular DL architectures, several impressive convolutional neural network (CNN) structures, such as ResNet [7] and U-Net [8], have already been widely explored for RS tasks. However, since the goal of CNNs is to learn a parametric translation function by using a dataset of input-output examples, considerable manual efforts are needed for designing effective losses between predicted and ground truth pixels. To address this problem, generative adversarial networks [9] were recently proposed, which learn a mapping from input to output images and tries to classify if the output image is real or fake.

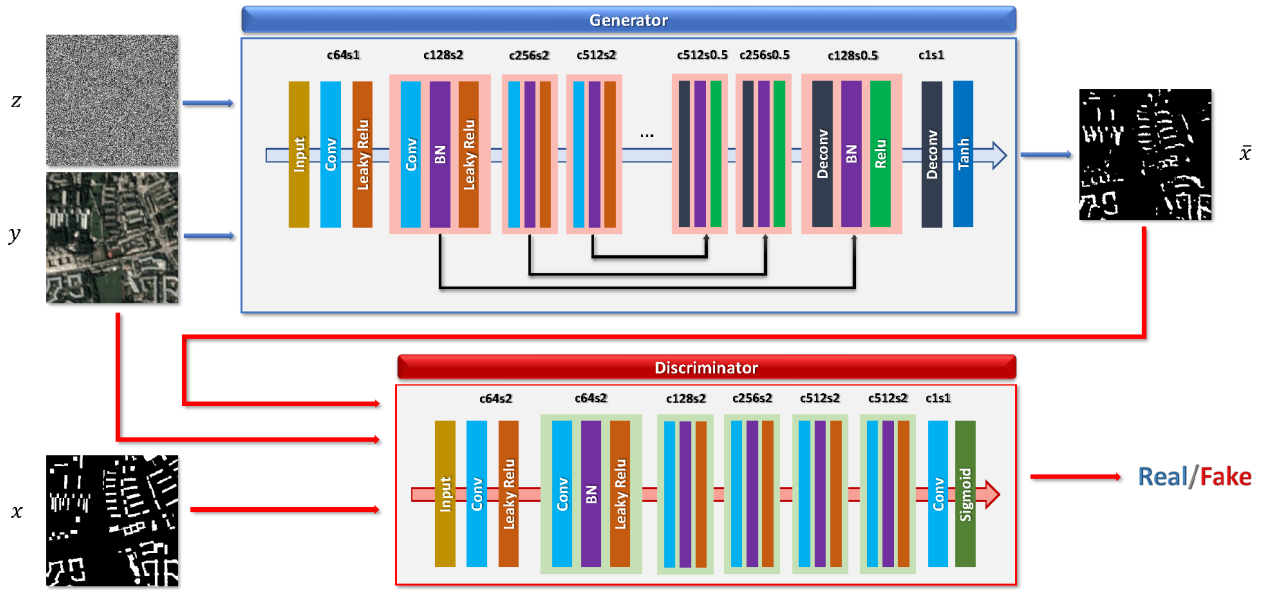

Fig. 2. Network architecture of the proposed method

In this regard, one of the motivations of this study was to explore the potential of GANs in building footprint generation by comparing their performance with other CNN structures. However, GANs also have their own limitations: (a) there is no control over the modes of data being generated, (b) and the training is delicate and unstable. Therefore, several studies have proposed alternatives to traditional GANs, such as conditional GANs [10] and Wasserstein GANs [11]. In order to direct the data generation process and improve the stability of training, we propose combining a conditional GAN, a Wasserstein GAN, and a gradient penalty term for building footprint generation, which are exploited for the first time in the remote-sensing community.

The proposed building footprint generation method is described in Section II. In Section III, the details of the datasets and the experimental results are presented and analyzed. The final conclusions follow in Section IV.

## II. METHODOLOGY

### A. Review of GANs

GANs were firstly proposed in [9] and consist of two neural networks: generator $G$ takes noise variables as input to generate new data instances while discriminator $D$ decides whether each instance of data belongs to the actual training dataset or not. $D$ and $G$ play a two-player minimax game with the objective function as

$$\mathcal{L}_{GAN} = E_{p_x}[\log D(x)] + E_{p_z}[\log(1 - D(G(z)))] \quad (1)$$

where $E$ is the empirical estimation of the expected value of the probability. $x$ is the training data with the true data distribution $p_x$, $z$ represents the noise variable sampled from distribution $p_z$, and $\bar{x} = G(z)$ represents the generated data instances. $G$ and $D$ are trained simultaneously: for $G$ to minimize $\log(1 - D(G(z)))$ and for $D$ to maximize $\log D(x)$.

To address the problem of no control over the modes of data being generated in GANs, Mirza et al. [10] extended GANs to a conditional model, where both the generator and discriminator are conditioned on certain extra information $y$, which could be any kind of auxiliary information, such as class labels. The conditioning is performed by feeding $y$ into both the discriminator and generator as an additional input layer. The objective function of CGANs is constructed as following:

$$\mathcal{L}_{CGAN} = E_{p_x}[\log D(x|y)] + E_{p_z}[\log(1 - D(G(z|y)))] \quad (2)$$

In order to improve the stability of learning of GANs and remove problems like mode collapse, WGANs were proposed by Arjovsky et al. [11], which use an alternative cost function that is derived from an approximation of the Wasserstein distance. They are more likely to provide gradients that are useful for updating the generator than the original GANs.

### B. Proposed Method

In this work, we want to exploit the superiorities of both CGANs and WGANs. Therefore, we propose CWGANs, which can impose a control on the modes of data being generated, and can also achieve more stable training as well. The objective function of CWGANs is given by:

$$\mathcal{L}_{CWGAN} = E_{p_x}[D(x|y)] - E_{p_z}[D(G(z|y))] \quad (3)$$

However, due to the use of weight clipping in WGANs, CWGANs may still generate low-quality samples or fail to converge in some settings. Therefore, we used an alternative to clipping weights: the addition of a gradient penalty term

[12] with respect to its input, whose objective function can be written as:

$$\mathcal{L}_{GP} = \lambda_1 E_{p_{x,z}}[(||\nabla D(\alpha x + (1-\alpha)G(z|y))||_2 - 1)^2] \quad (4)$$

where $\lambda_1$ is the gradient penalty coefficient, and $\alpha$ is a random number with uniform distribution in $[0, 1]$.

In order to let the generator to be located near the ground truth output and to decrease blurring, a traditional loss $L_1$ distance is mixed with the CWGAN objective:

$$\mathcal{L}_{L_1} = \lambda_2 E_{p_{x,z}}[||x - G(z|y)||_1] \quad (5)$$

where $\lambda_2$ is the coefficient for $L_1$ regularization. Finally, our objective function is the combination of CWGAN, gradient penalty term, and $L_1$ regularization.

$$\mathcal{L} = \arg\min_G \max_D \mathcal{L}_{CWGAN} + \mathcal{L}_{GP} + \mathcal{L}_{L_1} \quad (6)$$

*C. Network Architectures*

The network architecture in this work is shown in Fig. 2, which is used to generate the building footprint from satellite imagery.

We used the U-Net as the generator architecture. It is an encoder-decoder network with skip connections to concatenate all channels at layer $i$ with those at layer $n - i$, where $n$ is the total number of layers. The Leaky ReLU activation is used for the downsampling process, and the ReLU activation is used for upsampling. The aim of the encoder is to match the input and output into an embedded space while the decoder constrains the mapping spaces to allow a good reconstruction of the original input and output. Since skip connection can concatenate different layers, the U-Net can shuttle the low-level information (e.g., edges) directly across the net from input to output.

As for the discriminator architecture, the PatchGAN proposed in [13] is exploited to model a high frequency structure. This network tries to classify whether each patch in an image is real or fake. With the discriminator running convolutionally across the image, the ultimate output of $D$ can be provided by averaging all responses. The PatchGAN effectively models the image as a Markov random field and can therefore be understood as a form of texture.

III. EXPERIMENTS

*A. Description of Datasets*

In this work, we chose two study areas in Germany, which were Munich and Berlin. We used PlanetScope satellite imagery with three bands (R, G, B) and a spatial resolution of 3 m to test our proposed method. The corresponding building footprints were downloaded from OpenStreetMap (OSM). We processed the imagery using a $256 \times 256$ sliding window with a stride of 75 pixels to produce around 3000 sample patches. The sample patches were divided into two parts, where 70% were used to train the network and 30% were used to validate the trained model.

*B. Experimental Setup*

The number of both generator and discriminator filters in the first convolution layer was 64. The downsample factor is 2 in both the discriminator and the encoder of the generator. In the decoder of the generator, deconvolutions were performed with an upsample factor of 2. All convolutions and deconvolutions had a kernel size of $4 \times 4$, a stride equal to 2, and a padding size of 1. An Adam solver with a learning rate of 0.0002 was adopted as an optimizer for both networks. Furthermore, we use a batch size of one for each network and trained at 200 epochs. The clipping parameter in CWGAN was 0.01. For the CWGAN-GP, the gradient penalty coefficient $\lambda_1$ was set to 10 as recommended in [12]. Our networks were implemented with a Pytorch framework and trained on an NVIDIA TITAN X GPU with 12 GB of memory. Building footprint generation methods based on CGAN, U-Net, and ResNet-DUC in [14] were taken as the algorithms of comparison.

*C. Results and Analysis*

In this work, we evaluated the inference performances using metrics for a quantitative comparison: overall accuracy (OA), F1 scores, and IoU scores. Specifically, the F1 and IoU metrics are defined as follows:

$$\text{F1} = \frac{2 \times \text{precision} \times \text{recall}}{\text{precision} + \text{recall}} \quad (7)$$

$$\text{IoU} = \frac{\text{TP}}{\text{TP} + \text{FP} + \text{FN}}. \quad (8)$$

where *TP* is the number of true positives, *FP* is the number of false positives and *FN* is the number of false negatives. The impacts of hyperparameters have been investigated for our proposed methods. Firstly, the influence of different depths $d$ of the U-Net structure has been explored.

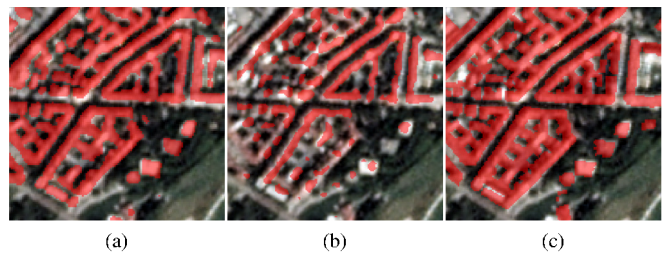

Fig. 3. Comparison of results generated by U-Net structure with different depths. (a) Depth ($d = 5$); (b) Depth ($d = 8$); (c) Ground truth.

Fig. 3 shows visual results of one patch with different depths compared to the ground truth. As one can see in Fig. 3, a large number of roofs are omitted by the network with $d = 8$ but are identified by the depth $d = 5$. Similar phenomenas have been reported in [15]. With the network depth increasing, accuracy gets saturated and then degrades rapidly, since adding more layers to a suitably deep model leads to a higher training error. Note that the optimal depth of the network should be comparable with the size of useful features in the imagery in order to achieve high accuracy.

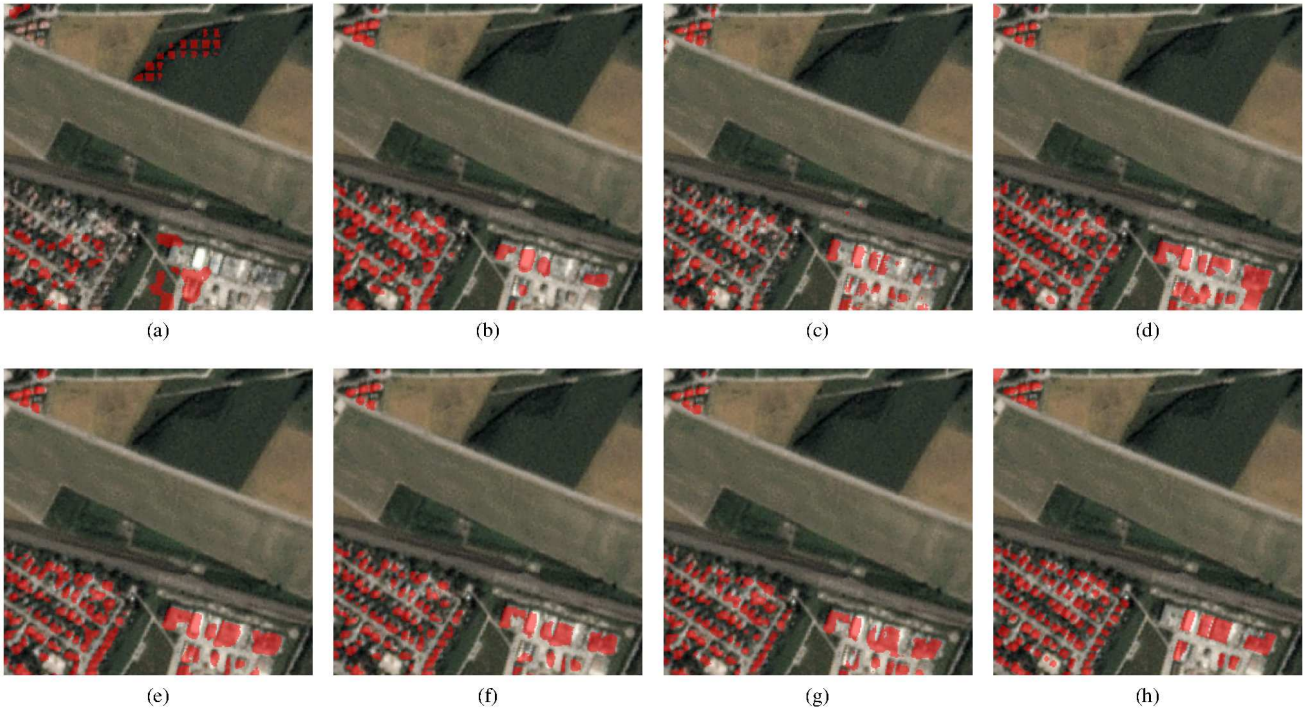

Fig. 4. Visualized comparison of different networks and coefficients $\lambda_2$ of $L_1$ loss; (a) CGAN ($\lambda_2 = 1$); (b) CGAN ($\lambda_2 = 100$); (c) ResNet-DUC; (d) U-Net; (e) CWGAN ($\lambda_2 = 100$); (f) CWGAN-GP ($\lambda_2 = 1$); (g) CWGAN-GP ($\lambda_2 = 100$); (h) ground truth.

Secondly, we have chosen different coefficients ($\lambda_2 = 1, 100$) of $L_1$ loss with the CGAN and CWGAN-GP. The quantitative results are listed in Table II, and results of the sample for visual comparison are in Fig. 4. And the comparison of training and inferencing time is shown in Fig. 5.

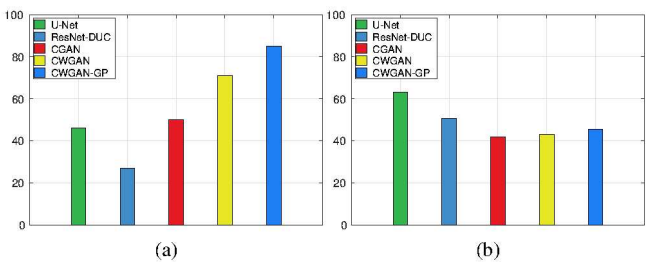

Fig. 5. Traning and inferencing time of different methods. (a) training time [s] (b) inferencing time [ms]

When the coefficient of $L_1$ loss increased from 1 to 100, the CGAN results dramatically improved for all evaluation metrics. As one can see from Fig. 4 (a) and (b), the building area generated by the CGAN with $\lambda_2 = 100$ is more correct and complete than that with $\lambda_2 = 1$. Such a result can be potentially explained by the fact that the $L_1$ loss term penalizes the distance between ground truth outputs and synthesized outputs, and the synthesized outputs from the $L_1$ loss term are better for the training of the discriminator. In contrast, the result of CWGAN-GP with $\lambda_2 = 100$ is slightly better than with $\lambda_2 = 1$, which indicates that our proposed method is not sensitive to hyperparameters. Moreover, it should be noted that the numerical results did not indicate a considerable difference when choosing different hyper-parameter combinations. This is due to the stability of our proposed methods, which nearly removes all hyperparameters tuning and simply uses the default setting.

TABLE I
COMPARISON OF DIFFERENT NETWORKS ON THE TEST DATASETS

| Methods | Overall Accuracy | F1 score | IoU score |
|---|---|---|---|
| CGAN ($\lambda_2 = 1$) | 79.42% | 0.1555 | 0.0842 |
| CGAN ($\lambda_2 = 100$) | 85.46% | 0.5787 | 0.4072 |
| ResNet-DUC | 85.71% | 0.5881 | 0.4166 |
| U-Net | 86.03% | 0.6455 | 0.4766 |
| CWGAN | 88.54% | 0.6737 | 0.5079 |
| CWGAN-GP ($\lambda_2 = 1$) | 88.87% | 0.6821 | 0.5169 |
| CWGAN-GP ($\lambda_2 = 100$) | **89.06%** | **0.6830** | **0.5194** |

Finally, we applied the selected coefficient of $L_1$ loss and depth ($d = 5$) in the generator to our proposed method CWGAN-GP. From Table I, we can see that the proposed method gives the best accuracy for all metrics. Compared to a CGAN, the CWGAN and CWGAN-GP indicate a dramatical increase of segmentation performance. This is because that even when two distributions are located in lower dimensional manifolds without overlaps, the Wasserstein distance can still provide a meaningful representation of the distance in-between. Since the weights in the discriminator of the CWGAN clamped to small values around zero, the parameters of the weights can lie in a compact space, which leads a learning process more stable than that of CGANs. However, a hyperparameter (the size of clipping window) in the CWGAN should still be tuned in order to avoid unstable training. If the clipping window is too large, there will be slow convergence after weight clipping. Moreover, if the clipping window is too small, it will lead to vanishing gradients. Therefore, the proposed CWGAN-GP, which add a gradient penalty term into the loss of discriminator, will improve the stability of the training. The proposed methods (CWGAN and CWGAN-GP) outperform ResNet-DUC in both numerical results and visual analysis, because the skip connections in generator $G$ combines both the lower and higher layers to generate the final output, retaining more details and better preserving the boundary of the building area. Compared to the U-Net, the proposed methods achieve higher overall accuracy, the F1 score and IoU score, as the min-max game between the generator and discriminator of the GAN, motivates both to improve their functionalities.

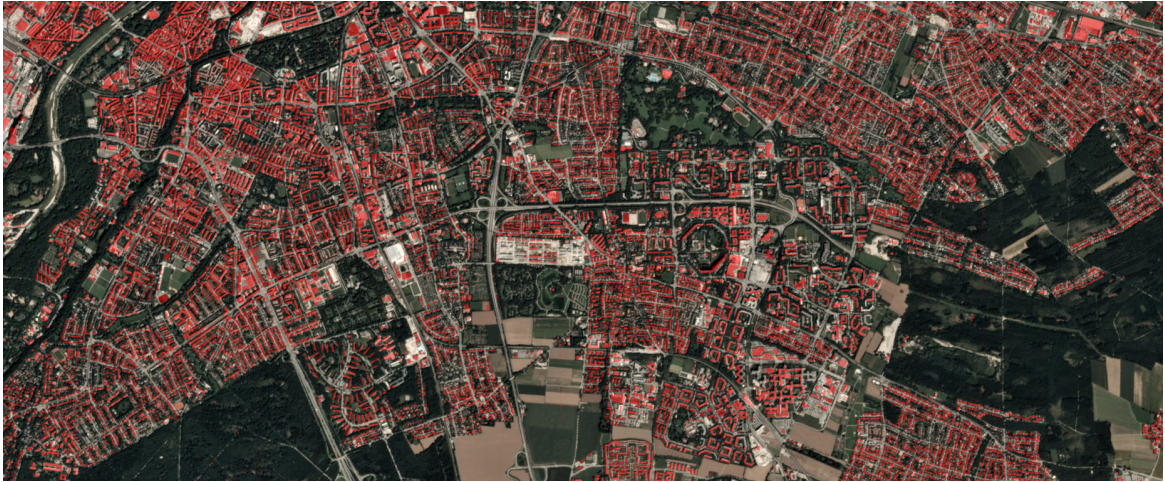

Fig. 6. Section of the entire Munich test area. Red color indicates the building footprint generated by the proposed method and overlays an optical image.

Fig. 6 presents a section of the entire Munich test area. The red color indicates the building footprint generated by the proposed method and overlays an optical image.

## IV. CONCLUSION

GANs, which have recently been proposed, provide a way to learn deep representations without extensively annotated training data. This research aimed to explore the potential of GANs in the performance of building footprint generation and improve its accuracy by modifying the objective function. Specifically, we proposed two novel network architectures (CWGAN and CWGAN-GP) that integrate CGAN and WGAN, as well as a gradient penalty term, which can direct the data generation process and improve the stability of training. The proposed method consists of two networks: (1) the U-Net architecture in the generator and (2) the PatchGAN in the discriminator. PlanetScope satellite imagery of Munich and Berlin was investigated to evaluate the capability of the proposed approaches. The experimental results confirm that the proposed methods can significantly improve the quality of building footprint generation compared to existing networks (e.g., CGAN, U-Net and ResNet-DUC). Additionally, it should be noted that the stability of our proposed method CWGAN-GP nearly removes all hyperparameters tuning.